\documentclass[conference]{IEEEtran}

\usepackage{cite}
\usepackage{amsmath,amssymb,amsfonts}
\usepackage{algorithmic}
\usepackage{graphicx}
\usepackage{textcomp}
\usepackage{xcolor}
\usepackage{multirow, multicol, makecell}
\usepackage{booktabs}
\usepackage{hyperref}
\usepackage{placeins}
\usepackage{float}
\usepackage{orcidlink}
\usepackage{xspace}

\usepackage[capitalize]{cleveref}
\crefname{section}{Sec.}{Secs.}
\Crefname{section}{Section}{Sections}
\Crefname{table}{Table}{Tables}
\crefname{table}{Tab.}{Tabs.}
\def\BibTeX{{\rm B\kern-.05em{\sc i\kern-.025em b}\kern-.08em
    T\kern-.1667em\lower.7ex\hbox{E}\kern-.125emX}}
\begin{document}

\title{\bmName: A Unified Benchmark for Federated Unsupervised Anomaly Detection in Tabular Data\\
}

\newcommand{\bmName}{FedAD-Bench\xspace}

\author{
Ahmed Anwar$^{1}$ \orcidlink{0009-0004-9737-7177},
Brian Moser$^{1,2}$ \orcidlink{0000-0002-0290-7904}
Dayananda Herurkar$^{1,2}$ \orcidlink{0000-0003-1556-8769},
Federico Raue$^{1}$ \orcidlink{0000-0002-8604-6207},\\
Vinit Hegiste$^{2}$ \orcidlink{0000-0001-6944-1988},
Tatjana Legler$^{1,2}$ \orcidlink{0000-0002-7297-0845},
and Andreas Dengel$^{1,2}$ \orcidlink{0000-0002-6100-8255} \\ \\
$^{1}$German Research Center for Artificial Intelligence (DFKI), Kaiserslautern, Germany \\
$^{2}$RPTU Kaiserslautern-Landau, Germany

}

\maketitle

\begin{abstract}
The emergence of federated learning (FL) presents a promising approach to leverage decentralized data while preserving privacy. 
Furthermore, the combination of FL and anomaly detection is particularly compelling because it allows for detecting rare and critical anomalies (usually also rare in locally gathered data) in sensitive data from multiple sources, such as cybersecurity and healthcare.
However, benchmarking the performance of anomaly detection methods in FL environments remains an underexplored area. 
This paper introduces \bmName, a unified benchmark for evaluating unsupervised anomaly detection algorithms within the context of FL. 
We systematically analyze and compare the performance of recent deep learning anomaly detection models under federated settings, which were typically assessed solely in centralized settings. 
\bmName encompasses diverse datasets and metrics to provide a holistic evaluation. 
Through extensive experiments, we identify key challenges such as model aggregation inefficiencies and metric unreliability. 
We present insights into FL's regularization effects, revealing scenarios in which it outperforms centralized approaches due to its inherent ability to mitigate overfitting. 
Our work aims to establish a standardized benchmark to guide future research and development in federated anomaly detection, promoting reproducibility and fair comparison across studies.
\end{abstract}

\begin{IEEEkeywords}
Federated Learning, Distributed Machine Learning, Unsupervised Anomaly Detection, Outlier Detection, Benchmark
\end{IEEEkeywords}

\section{Introduction}
Recently, FL \cite{federatedlearning17} has been growing in interest rapidly due to its appealing properties, especially in privacy-preserving machine learning. 
The main concept of FL is the aggregation of \textit{local} models, which is used to extract a \textit{global} model that matches the performance of centralized models while ensuring that the training data remains confidential (local).
This straightforward yet powerful approach has demonstrated its effectiveness across a range of applications and fields, such as Text Prediction \cite{googlekeyboard}, Internet of Things \cite{diot}, Image Super-Resolution \cite{flsr}, Manufacturing \cite{hegiste2022fedimg} and Healthcare \cite{patientclusteringFL}. 

On the other hand, outlier detection remains a critical challenge in data analysis, where it is essential to identify samples that deviate from expected and known distributions \cite{deepADReview}. 
While tree-based methods, such as XGBoost \cite{xgboost} and RecForest \cite{recforest}, achieve state-of-the-art performance on outlier detection datasets \cite{isitworthit, cantreebasedsurpassdeep}, unsupervised outlier detection has also proven to be highly effective, and its performance is on par with tree-based methods \cite{alvarez2022revealing}. 
Additionally, anomaly detection data, such as network intrusion and medical records, is often distributed among organizations that wish to keep it private. 
FL would allow for these organizations to collaboratively train high-performing machine learning models while preserving privacy and abiding by data protection regulations, which significantly benefits all participants \cite{finfedod}. 
Unfortunately, there is a significant lack of evaluation of anomaly detection methods in FL settings.
Furthermore, a convention for testing such methods in FL settings is yet to be established.

In response, we introduce \bmName, a unified benchmark designed to evaluate state-of-the-art unsupervised models for anomaly detection in FL settings. 
Our focus is specifically on deep learning for several compelling reasons:

Firstly, while tree-based methods are often the most effective for anomaly detection, their integration with FL presents significant challenges. 
These methods require specialized aggregation techniques, complicating their implementation in FL environments \cite{secureboost}. 
Secondly, deep learning methods have predominantly been applied to image and natural language processing tasks, leaving a notable gap in applying deep anomaly detection techniques to tabular data within FL frameworks. This work addresses this gap by introducing a consistent framework for evaluating deep anomaly detection algorithms in FL.

We established multiple conventions and features in \bmName, inspired by benchmarks in centralized settings, such as those by Alvarez et al. \cite{alvarez2022revealing}. Key aspects of \bmName include:
\begin{itemize}
\item \textbf{Support for Federated Learning.} \bmName is tailored to evaluate anomaly detection methods within federated settings, ensuring relevance to real-world FL scenarios. We provide practical recommendations for evaluating future methods and offer a robust codebase to promote reproducibility and applicability.

\item \textbf{Redesigned Data Splitting.} In traditional centralized benchmarks, data splitting often includes anomalies in the training set. However, since unsupervised methods aim to learn the underlying distributions or features of normally distributed data, we find it more appropriate to exclude anomalies from the training set. This redesign enhances the model's ability to learn normal data characteristics without interference.
\item \textbf{Unified Set of Evaluation Metrics for Robust Results.} Anomaly detection metrics are often threshold-dependent, which can lead to manipulation and misleading results. To counter this, we propose a data-splitting approach that places all anomalies in the test set, reducing bias. Moreover, \bmName employs a robust set of evaluation metrics, including Precision, Recall, AUROC, AUPR, and F1 Score, providing a comprehensive assessment of model performance
\end{itemize}

By introducing these conventions and features, \bmName sets a standardized benchmark for testing new and emerging methods on tabular data in FL settings, promoting reproducibility and fair comparison across studies. 
This benchmarking framework aims to guide future research and development in federated anomaly detection, ensuring that advancements are grounded in rigorous and consistent evaluation practices.

\begin{figure*}[!t]
  \centering
  \includegraphics[width=.75\linewidth]{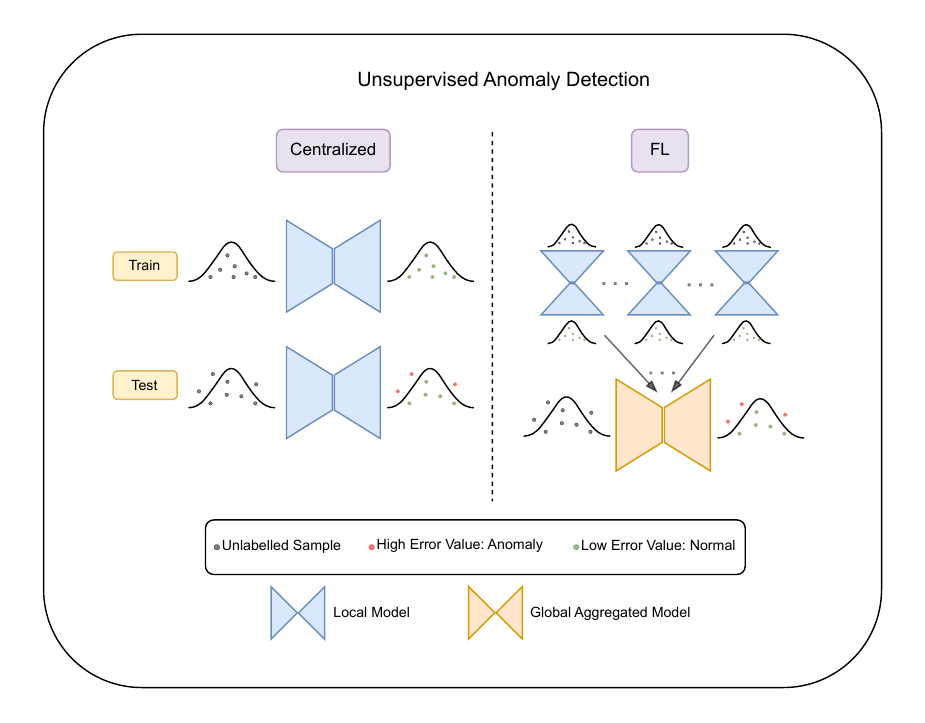}
  \caption{\label{fig:main}Unsupervised anomaly detection setup in both centralized and FL scenarios. 
  Training is done on 50\% of the normal data while the test set contains all anomalies as well as the other half of the normal data. Each client trains their local model on their own set of inliers. Afterward, the trained models are aggregated at the server using FedAvg and evaluated on the test set. The inference is based on the reconstruction error, where samples with high error values are considered anomalies. 
}
\end{figure*}

\section{Related Work}\label{sec:RelatedWork}
The intersection of FL and anomaly detection represents an emerging area of research with significant potential for enhancing data privacy and security across various domains. This section reviews relevant literature, focusing on unsupervised methods for anomaly detection and the application of FL in this context. The aim is to highlight existing approaches and underscore the need for standardized benchmarks in evaluating these techniques.
\subsection{Unsupervised Methods for Anomaly Detection}
Unsupervised learning for anomaly detection depends on learning inlier (normal) data distribution and finding a threshold such that a sample distance from this distribution, measured using a certain distance metric, indicates whether it belongs to it or is an outlier (abnormal) sample \cite{boostf1}. To this end, many distance metrics are used in the literature \cite{deepADReview}. 
Yet, we will mention the methods from which we chose representatives for brevity. Reconstruction methods such as Memory Augmented Deep Auto Encoder (MemAE) \cite{memae} and Deep Auto Encoder (DAE) \cite{dae} are among the most famous methods for unsupervised anomaly detection. They are trained to reconstruct normal samples and are used to classify anomalies according to the reconstruction loss. One-class classification methods such as Deep Support Vector Data Description (DeepSVDD) \cite{deepsvdd} learn a description of the normal data through training and recognize abnormal samples as the ones that do not follow the same description. Methods such as Deep Auto Encoding Gaussian Mixture Model (DAGMM) \cite{dagmm} learn the probability density function of the normal data to achieve the classification of abnormal data similar to previous methods. Furthermore, the Deep Structured Energy Based Model (DSEBM) \cite{dsebm} uses both energy as well as reconstruction scores as criteria for anomaly detection. Finally, Neural Transformation Learning for Deep Anomaly Detection (NeuTraLAD) \cite{neutralad} uses deterministic contrastive learning and learnable transformations. NeuTraLAD consists of a neural transformation module and an encoder, which are both trained using deterministic contrastive loss (DCL). The DCL is used during inference to calculate an anomaly score to denote a sample as normal or abnormal.
\subsection{Federated Learning}
FL has been increasingly explored in the context of anomaly detection, with numerous studies investigating their combined application. IoT has seen the most attention towards FL and anomaly detection, D{\"I}OT is a self-supervised approach for detecting compromised IoT devices, especially by intrusion attacks \cite{diot}. This framework uses FL to aggregate IoT devices' behavior profiles to detect anomalies. Li et al. \cite{anomalyweightupdate} utilized a pre-trained anomaly detection model on the server side to detect anomalous weight updates propagated by the clients. Rey et al. \cite{iotmalwareFL} used supervised and unsupervised models to present a use case for malware detection in IoT devices. They further focus on adversarial attacks and communication costs. Wang et al. \cite{wangFLADcase} used FL to train a global anomaly detection model aiming at finding battery failures in battery-backed energy storage systems. Lastly, Herurkar et al. \cite{finfedod} applied representation learning and FL to create Fin-Fed-OD, an algorithm that distinguishes between distinct outliers across different clients in FL. While these efforts concentrate on specific application domains, our study aims to establish a standardized framework for testing and benchmarking such approaches, providing a consistent basis for evaluating the effectiveness of FL in anomaly detection.   

\section{\bmName }
\label{sec:Methodology}

To establish a robust benchmark for unsupervised anomaly detection in FL environments, we designed a comprehensive evaluation framework called \bmName. 
Our benchmark is inspired by recent advancements in both anomaly detection \cite{alvarez2022revealing} and FL, aiming to bridge the gap between these two domains. The primary components of \bmName can be categorized into data splitting, evaluation metrics, and federated learning, which are detailed below.

We will use \bmName in our experimental section to give initial results for new anomaly detection methods specifically designed for FL.
Moreover, it can be used to evaluate new centralized solutions directly in FL settings.
Overall, \bmName aims to provide a standardized and reproducible framework for evaluating unsupervised anomaly detection models in FL settings, guiding future research and promoting fair comparisons across studies.

\subsection{Data Splitting}
\label{sec:data_splitting}
While it is common in deep learning for classification to do 80-20 (train-test) and 70-10-20 (train-validation-test) splits of the dataset, in unsupervised anomaly detection, it is more beneficial to do a class-based split \cite{alvarez2022revealing}. Given that the aim of unsupervised methods is to learn underlying distributions or features of \textit{normally distributed} data, removing anomalies from the training set is thus more appropriate. 
Therefore, we advocate for test sets that include all anomalies in the data and half the number of inliers. 
This approach ensures a more accurate assessment of the model's ability to detect anomalies in a realistic setting, and sets a standard for data splitting for a fair comparison between different methods.

\subsection{Evaluation Metrics}
 One main issue with metrics used for anomaly detection evaluation is that they are threshold dependent, such as the F1-score; this dependency means that they can be easily manipulated by fixing the threshold and changing the number of anomalies in the test set to achieve better results \cite{boostf1}. 
 The data-splitting approach we described earlier helps mitigate this potential bias by putting all anomalies into the test set. 
 In our experiments, we use the optimal thresholds for each trained model for a fair comparison and report Precision, Recall, the area under the ROC curve (AUROC), the area under the Precision-Recall curve (AUPR), and the F1 Score. 
 Estimating the optimal threshold involves iterating through various threshold values, calculating performance metrics (precision, recall, F1-Score) for each threshold, and identifying the threshold that maximizes the F1-Score. 
 We propose splitting the test set further into validation and test sets, the purpose of the validation set is to calculate the optimal threshold to be used for threshold-based metrics. 
 Moreover, this validation set should have the same anomaly ratio as the test set (a \textit{stratified} split).
 The threshold range is adjusted based on the fraction of normal data in the validation set and uses these metrics to ensure the threshold selected offers the best trade-off between precision and recall \cite{alvarez2022revealing}.

\subsection{Federated Learning}
In our study, we employ Federated Averaging (FedAvg) \cite{federatedlearning17} as the aggregation algorithm for our main FL experiments. Afterwards, we conduct extensive experiments using FedProx \cite{fedprox} to demonstrate the importance of aggregation strategies as one of the main axes of FL experiments in \Cref{subsec:fedprox}.

Our primary FL configuration involves three clients, though we also examine scenarios with five and seven clients to assess scalability. As described in \autoref{sec:data_splitting}, we exclude all anomalies from the training sets, incorporating them only in the test set. 
Consequently, the class imbalance problems in FL are beyond the scope of this paper, and we conduct experiments with a uniform sample distribution among the clients.

\section{Experiments}\label{sec:Experiments}
With \bmName, we also establish a strong baseline for future methods to test against. 
As such, we describe the datasets we used and the evaluated (existing) anomaly detection methods adapted to work within the FL setting in the following.

\begin{table}[t]
    \centering
    \caption{\label{tab:trainingsetup} Model parameters used for all experiments.}
    \begin{tabular}{l  l r c ccc}
    \toprule
    {\rotatebox[origin=c]{90}{DS}} & Model  & Latent Dim. & \multicolumn{3}{l}{Model Specific}\\
     \midrule
     \multirow{5}{*}{\rotatebox[origin=c]{90}{Arrythmia}} & DAE  & 3   &&& \\
      & DSEBM  & 2  &&& \\
      & DEEPSVDD & - & \multicolumn{3}{l}{\#Output Features: 64} \\
      & NeuTraLAD  & 32  & \multicolumn{3}{l}{Trans. Type: residual} \\
     & MemAE  & 3  & \multicolumn{3}{l}{Memory Dimension: 50} \\
     \midrule
      \multirow{5}{*}{\rotatebox[origin=c]{90}{Thyroid}}
        & DAE  & 2  &&& \\
      & DSEBM  & 2  &&& \\
      & DEEPSVDD & - & \multicolumn{3}{l}{\#Output Features: 1} \\
      & NeuTraLAD  & 24  & \multicolumn{3}{l}{Trans. Type: residual} \\
     & MemAE  & 3  & \multicolumn{3}{l}{Memory Dimension: 50} \\
     \midrule
     \multirow{5}{*}{\rotatebox[origin=c]{90}{KDD10}}  
     & DAE  & 2  &&& \\
      & DSEBM  & 512  &&& \\
      & DEEPSVDD & - & \multicolumn{3}{l}{\#Output Features: 29} \\
      & NeuTraLAD  & 32  & \multicolumn{3}{l}{Trans. Type: multiplicative} \\
     & MemAE  & 3  & \multicolumn{3}{l}{Memory Dimension: 50} \\
     \midrule
     \multirow{5}{*}{\rotatebox[origin=c]{90}{NSL-KDD}} 
     & DAE  & 2  &&& \\
      & DSEBM  & 512  &&& \\
      & DEEPSVDD & - & \multicolumn{3}{l}{\#Output Features: 31} \\
      & NeuTraLAD  & 32  & \multicolumn{3}{l}{Trans. Type: multiplicative} \\
     & MemAE  & 3  & \multicolumn{3}{l}{Memory Dimension: 50} \\
     \bottomrule
    \end{tabular}
\end{table}

\subsection{Datasets}
We chose 2 datasets from the medical domain and another 2 from the cybersecurity domain since these are among the most popular domains for tabular anomaly detection \cite{alvarez2022revealing}: 
\begin{itemize}
    \item Arrythmia \cite{arrhythmia}: a medical dataset detecting cardiac arrhythmia and originally classify it in one of the 16 groups. The classes are combined into presence or absence as a binary classification. The dataset contains 452 samples with 274 features, with an anomaly sample ratio of 0.1460.
    \item Thyroid \cite{thyroid}: another medical dataset with one anomalous class and two normal classes, which are combined into one. The anomaly ratio is the lowest in the collection of datasets, with a value of 0.0246 among 3772 samples with 6 features each.  
    \item KDDCUP10: this is a 10\% subset of the KDDCUP \cite{kdd} network intrusion detection dataset with 494021 samples characterized by 34 continuous and 7 categorical variables. Originally consisting of 5 classes, including 4 different attack types, all attacks are merged into a single anomaly class resulting in an anomaly ratio of 0.1969.
    \item NSL-KDD \cite{nslkdd}: this is considered an updated version of the KDDCUP dataset consisting of the same feature set and 148517 samples. The anomaly ratio, however, is much higher at 0.4811.
   
\end{itemize}

\subsection{Models}
As per our selection of models, we chose 5 state of the art models that achieved the highest score on at least one evaluation metric and one dataset \cite{alvarez2022revealing}. This decision resulted in the following models, which we mentioned previously in \Cref{sec:RelatedWork}:
\begin{itemize}
    \item DAE \cite{dae}: learns to reconstruct normal data patterns. The model is trained on inliers, and identifies anomalies by measuring the reconstruction error, such that if the error is high, the data point is considered anomalous.
    \item DSEBM \cite{dsebm}: uses an energy-based formulation to model the distribution of normal data. The energy function is learned by the model, and anomalies are detected based on their energy values, where higher energy indicates an anomaly.
    \item DEEPSVDD \cite{deepsvdd}: is built on top of kernel-based one-class classification. It utilizes a neural network to map data into a hypersphere in a high-dimensional space, and anomalies are identified as data points that lie outside the hypersphere. The model aims at minimizing the volume of the hypersphere containing normal data.
    \item NeuTraLAD\cite{neutralad}: leverages neural networks to learn transformations that enhance the separation between normal and anomalous data. Learning optimal transformations makes anomalies more distinguishable from normal data, thereby improving detection accuracy.
    \item MemAE\cite{memae}: enhances the traditional autoencoder by incorporating a memory module. This module stores prototypes of normal patterns, allowing the model to reconstruct normal data better and detect anomalies through reconstruction errors. The memory module helps in handling diverse and complex data patterns effectively.
\end{itemize}

\subsection{Hyper-Parameters}

For experiments Adam \cite{adam} optimizer was used with a learning rate and weight decay value of $1e-4$. 
The batch size for centralized experiments was 128 samples for Arrythmia and Thyroid datasets, and 1024 for KDD10 and NSL-KDD datasets.
In FL setting, the same batch size values were used as long as the local partitions were big enough, otherwise, the batch size was decreased to a suitable value.
Each model was trained for the respective optimal number of epochs to establish a legitimate comparison between the performances of the models, the optimal number of epochs follows the results of Alvarez et al. benchmark \cite{alvarez2022revealing}.
The number of local epochs for FL clients is set to 10, and the number of communication rounds is equal to $E/10$ where $E$ is the optimal number of epochs in the centralized scenario.
For some experiments such as NeutraLAD and MemAE on KDD10 dataset the number of communication rounds $E/10$ was too few for them to converge and was set to $E$ instead. 

\section{Results}

\begin{table*}[t!]
    \centering
    \caption{\label{tab:main_results}Results of 5 state-of-the-art models on 4 datasets in both centralized and 3-client FL scenarios.}
    \begin{tabular}{l  l c cc c cc c cc c cc c cc}
    \toprule
     \multirow{2}{*}{\rotatebox[origin=c]{90}{DS}}  & \multirow{2}{*}{Method} && \multicolumn{2}{c}{Precision} && \multicolumn{2}{c}{Recall} && \multicolumn{2}{c}{AUROC} && \multicolumn{2}{c}{AUPR}  && \multicolumn{2}{c}{F1}\\
     &&& Centralized & FL && Centralized & FL && Centralized & FL && Centralized & FL && Centralized & FL \\
     \midrule
     \multirow{5}{*}{\rotatebox[origin=c]{90}{Arrythmia}} 
        & DAE && \textbf{0.70} & 0.65 && \textbf{0.54} & \textbf{0.54} && \textbf{0.85} & 0.81 && \textbf{0.68} & 0.67 && \textbf{0.61} & 0.59 \\
        & DSEBM && \textbf{0.58} & 0.56 && 0.57 & \textbf{0.63} && \textbf{0.82} & 0.81 && \textbf{0.69} & 0.66 && 0.57 & \textbf{0.59} \\
        & DEEPSVDD && 0.61 & \textbf{0.62} && \textbf{0.38} & \textbf{0.38} && 0.57 & \textbf{0.69} && 0.43 & \textbf{0.55} && 0.47 & \textbf{0.48} \\
        & NeuTraLAD && \textbf{0.67} & 0.66 && \textbf{0.42} & 0.40 && \textbf{0.74} & \textbf{0.74} && \textbf{0.59} & 0.58 && \textbf{0.52 }& 0.50 \\
        & MemAE && 0.71 & \textbf{0.72} && 0.38 & \textbf{0.40} && \textbf{0.81} & \textbf{0.81} && \textbf{0.66} & \textbf{0.66} && 0.50 & \textbf{0.52} \\
    \midrule
      \multirow{5}{*}{\rotatebox[origin=c]{90}{Thyroid}} 
        & DAE && \textbf{0.22} & \textbf{0.22} && \textbf{0.43} & \textbf{0.43} && \textbf{0.86} &\textbf{ 0.86} && \textbf{0.28} & \textbf{0.28} && \textbf{0.29} & \textbf{0.29} \\
        & DSEBM && \textbf{0.22} & 0.20 && \textbf{0.16} & 0.15 && \textbf{0.74} & 0.73 && \textbf{0.20} & 0.18 && \textbf{0.19} & 0.17 \\
        & DEEPSVDD && 0.23 & \textbf{0.42} && \textbf{0.51} & 0.27 && 0.71 & \textbf{0.76 }&& \textbf{0.35} & 0.26 && 0.32 & \textbf{0.33} \\
        & NeuTraLAD && \textbf{0.68} & 0.57 && 0.68 & \textbf{0.80} && 0.93 & \textbf{0.96} && 0.61 & \textbf{0.68} && \textbf{0.68} & 0.66 \\
        & MemAE && \textbf{0.22} & \textbf{0.22} && \textbf{0.43} & \textbf{0.43} && \textbf{0.86} & \textbf{0.86} && \textbf{0.28} & \textbf{0.28} && \textbf{0.29 }& \textbf{0.29} \\
    \midrule
      \multirow{5}{*}{\rotatebox[origin=c]{90}{KDD10}} 
        & DAE && \textbf{0.93} & \textbf{0.93} && \textbf{1.00} &  \textbf{1.00} & & \textbf{0.98} & \textbf{0.98}&& \textbf{0.94} & \textbf{0.94} && \textbf{0.96} & \textbf{0.96} \\
        & DSEBM && \textbf{0.97} & \textbf{0.97} && \textbf{1.00} &  0.99 && \textbf{0.99} & \textbf{0.99} && 0.96 & \textbf{0.97} && \textbf{0.98} & \textbf{0.98} \\
        & DEEPSVDD && \textbf{0.88} & 0.73 && \textbf{0.99} & 0.83 && \textbf{0.99} & 0.87 && \textbf{0.97} & 0.86 && \textbf{0.93} & 0.78 \\
        & NeuTraLAD && 0.57 & \textbf{0.60} && 0.50 & \textbf{0.69} && 0.62 & \textbf{0.65} && \textbf{0.66} & 0.56 && 0.53 & \textbf{0.64} \\
        & MemAE && \textbf{0.93} & 0.92 && \textbf{1.00} & 0.99 && \textbf{0.98} & \textbf{0.98} && \textbf{0.96} & 0.95 && \textbf{0.96} & 0.95 \\
    \midrule
      \multirow{5}{*}{\rotatebox[origin=c]{90}{NSL-KDD}} 
        & DAE && \textbf{0.96} & 0.93 && \textbf{0.94} & \textbf{0.94} && \textbf{0.98} & 0.97 && \textbf{0.99} & 0.98 && \textbf{0.95} & 0.93 \\
        & DSEBM && 0.94 & \textbf{0.95} && \textbf{0.95} & 0.92 && \textbf{0.98} & 0.97 && \textbf{0.99} & \textbf{0.99} && \textbf{0.94} & 0.93 \\
        & DEEPSVDD && \textbf{0.94} & 0.90 && \textbf{0.94} & 0.84 && \textbf{0.98} & 0.89 && \textbf{0.99} & 0.95 && \textbf{0.94 }& 0.87 \\
        & NeuTraLAD && 0.53 & \textbf{0.86} && 0.56 & \textbf{0.90} && 0.37 & \textbf{0.86} && 0.64 & \textbf{0.90} && 0.54 & \textbf{0.88} \\
        & MemAE && \textbf{0.94} & 0.92 && \textbf{0.95} & \textbf{0.95} && \textbf{0.98} & 0.97 && \textbf{0.99} & 0.98 && \textbf{0.94} & 0.93 \\
      \bottomrule
    \end{tabular}
\end{table*}

While it is expected for centralized learning to outperform FL, and this is the trend that is clear in \Cref{tab:main_results} where centralized learning outperforms FL in 8 out of the 20 experiments. However, FL runs also outperformed centralized in 4 cases such as NeuTraLAD on both KDD10 and NSL-KDD datasets. We consider a method to outperform the other if it performs better in at least one metric while not performing worse in any. The improvement in this case of FL results compared to centralized learning can be attributed to a better regularization of the models. Since after each FL round, the local models are all shifted to the global average model which can have positive effect on robustness, generalization, and can counter overfitting \cite{tatjanaFLreg}. 

Moreover, we see that certain models do not experience a performance drop when aggregated and can even outperform their centralized versions. DAE is a clear example of such a model, where it consistently performed better or at least on par with its centralized version. We conclude that its simple architecture is affected the least by the global aggregation, while benefiting from the regularization of FL in its generalization ability. Conversely, we find complex models such as DSEBM performing slightly worse in most of the cases. Such models need more specialised aggregation strategies to handle extra modules and heuristics which are part of the models' training.

It is important, to note that these results show how there can be a great discrepancy in conclusions depending on the metric that is looked at, which further demonstrates the importance of looking at all metrics holistically. Clear examples are results on the Thyroid dataset, where most models suffer poor performance on almost all metrics. Taking the DSEBM model as an example, in both centralized and FL experiments the model's score on each metric does not exceed 0.22 except for AUROC. On the other hand, when we only look at the AUROC metric, we see 0.74 and 0.73 scores for centralized and FL respectively. This can be misleading due to the fact that AUROC provides an extremely optimistic view by giving both normal and abnormal samples equal weights \cite{alvarez2022revealing}. We discuss this further in \Cref{subsec:unreliability}.

\subsection{Number Of Clients}\label{subsec:numclients}
We further analyze the effect of the number of active clients on the performance of unsupervised anomaly detection models in FL. The same experiments were executed on 5 and 7 clients, having more clients with the same dataset size means that each client will have less samples while fixing the number of local epochs as well as the number of FL communication rounds \cite{vucovich2022anomalydetectionfederatedlearning}. Consequently, models tend to take longer to converge with such small training dataset sizes, especially since the local gradients before aggregation point further away from the global optimum as the dataset size decreases. 
This is indeed apparent in \Cref{tab:FLclients} where in most scenarios the performance drops as the number of clients increase under the same number of local epochs and communication rounds.
Thyroid dataset saw the biggest drop in performance especially when using DEEPSVDD, on the other hand, performance of the models on KDD10 decreased the least.
The sustained performance on KDD10 dataset indicates that the samples are too homogeneous (not diverse enough) such that the local models are able to reach similar local minima on smaller number of samples

\begin{table*}[t!]
    \centering
    \caption{\label{tab:FLclients}Comparison between the results of FL incorporating 3, 5 and 7 clients.}
    \begin{tabular}{l  l c ccc c ccc c ccc c ccc c ccc}
    \toprule
     \multirow{2}{*}{\rotatebox[origin=c]{90}{DS}}  & Method && \multicolumn{3}{c}{Precision} && \multicolumn{3}{c}{Recall} && \multicolumn{3}{c}{AUROC} && \multicolumn{3}{c}{AUPR}  && \multicolumn{3}{c}{F1}\\
     & \# Clients && 3 & 5 & 7 && 3 & 5 & 7 && 3 & 5 & 7 && 3 & 5 & 7 && 3 & 5 & 7 \\
     \midrule
     \multirow{5}{*}{\rotatebox[origin=c]{90}{Arrythmia}} 
        & DAE && 0.65 & \textbf{0.76} & 0.67 && \textbf{0.54} & 0.42 & 0.50 && 0.81 & \textbf{0.83} & 0.82 && 0.67 & \textbf{0.69} & 0.67 && \textbf{0.59} & 0.54 & 0.57 \\
        & DSEBM && 0.0 & 0.57 & 0.61 && 0.0 & 0.63 & 0.53 && 0.0 & 0.81 & 0.81 && 0.0 & 0.65 & 0.67 && 0.0 & 0.6 & 0.57 \\
        & DEEPSVDD && \textbf{0.62} & 0.56 & 0.49 && \textbf{0.38} & \textbf{0.38} & 0.37 && \textbf{0.69} & 0.65 & 0.68 && 0\textbf{.55} & 0.51 & 0.54 && \textbf{0.48} & 0.45 & 0.42 \\
        & NeuTraLAD && \textbf{0.66} & 0.57 & 0.52 && 0.40 & 0.38 & \textbf{0.44} && \textbf{0.74} & 0.73 & 0.61 && \textbf{0.58} & 0.57 & 0.52 && \textbf{0.50} & 0.46 & 0.48 \\
        & MemAE && \textbf{0.72} & 0.71 & 0.71 && \textbf{0.40} & 0.38 & 0.38 && \textbf{0.81} & \textbf{0.81} & \textbf{0.81} && \textbf{0.66} & \textbf{0.66} & \textbf{0.66} && \textbf{0.52} & 0.50 & 0.50 \\
    \midrule
      \multirow{5}{*}{\rotatebox[origin=c]{90}{Thyroid}} 
        & DAE && \textbf{0.22} & \textbf{0.22} & \textbf{0.22} && \textbf{0.43} & \textbf{0.43} & \textbf{0.43} && \textbf{0.86} & \textbf{0.86} & \textbf{0.86} && \textbf{0.28} & \textbf{0.28} & 0.27 && \textbf{0.29} & \textbf{0.29} & \textbf{0.29} \\
        & DSEBM && 0.20 & 0.20 & \textbf{0.22} && 0.15 & 0.15 & \textbf{0.16} && \textbf{0.73} & \textbf{0.73} & \textbf{0.73} && 0.18 & 0.18 & \textbf{0.19} && 0.17 & 0.17 & \textbf{0.19} \\
        & DEEPSVDD && \textbf{0.42} & 0.08 & 0.09 && \textbf{0.27} & 0.15 & 0.15 && \textbf{0.76} & 0.54 & 0.58 && \textbf{0.26} & 0.07 & 0.08 && \textbf{0.33} & 0.11 & 0.11 \\
        & NeuTraLAD && \textbf{0.57} & 0.38 & 0.15 && \textbf{0.80} & 0.68 & 0.35 && \textbf{0.96} & 0.91 & 0.79 && \textbf{0.68} & 0.55 & 0.13 && \textbf{0.66} & 0.49 & 0.21 \\
        & MemAE && \textbf{0.22} & \textbf{0.22} & \textbf{0.22} && \textbf{0.43} & \textbf{0.43} & \textbf{0.43} && \textbf{0.86} & \textbf{0.86} & \textbf{0.86} && \textbf{0.28} & \textbf{0.28} & 0.27 && \textbf{0.29} & \textbf{0.29} & \textbf{0.29} \\
    \midrule
      \multirow{5}{*}{\rotatebox[origin=c]{90}{KDD10}} 
        & DAE && 0.93 & 0.93 & \textbf{0.96} && \textbf{1.00} &  \textbf{1.00} &  \textbf{1.00} & & 0.98 & 0.98 & \textbf{0.99} && 0.94 & 0.92 & \textbf{0.95} && 0.96 & 0.96 & \textbf{0.98} \\
        & DSEBM && \textbf{0.97} & \textbf{0.97} & \textbf{0.97} && \textbf{0.99} & \textbf{0.99} & \textbf{0.99} && \textbf{0.99} & \textbf{0.99} & \textbf{0.99} && \textbf{0.97} & \textbf{0.97} & \textbf{0.97} && \textbf{0.98} & \textbf{0.98} & \textbf{0.98} \\
        & DEEPSVDD && 0.73 & 0.81 & \textbf{0.83} && \textbf{0.83} & 0.68 & \textbf{0.83} && \textbf{0.87} & 0.73 & \textbf{0.87} && \textbf{0.86} & 0.78 & \textbf{0.86} && 0.78 & 0.74 & \textbf{0.83} \\
        & NeuTraLAD && 0.60 & \textbf{0.81} & 0.60 && \textbf{0.69} & 0.68 & 0.68 && 0.65 & 0.76 & \textbf{0.80}&& 0.56 & \textbf{0.77} & 0.51 && 0.64 & \textbf{0.74} & 0.64 \\
        & MemAE && 0.92 & 0.92 & \textbf{0.93} && 0.99 & \textbf{1.00} &  0.99 && \textbf{0.98} & \textbf{0.98} & \textbf{0.98} && \textbf{0.95} & \textbf{0.95} & \textbf{0.95} && 0.95 & \textbf{0.96} & \textbf{0.96} \\
    \midrule
      \multirow{5}{*}{\rotatebox[origin=c]{90}{NSL-KDD}} 
        & DAE && \textbf{0.93} & 0.92 & \textbf{0.93} && \textbf{0.94} & 0.93 & 0.92 && \textbf{0.97} & 0.96 & 0.96 && \textbf{0.98} & \textbf{0.98} & \textbf{0.98} && \textbf{0.93} & \textbf{0.93} & \textbf{0.93} \\
        & DSEBM && \textbf{0.95} & 0.90 & 0.90 && 0.92 & \textbf{0.96} & \textbf{0.96} && \textbf{0.97} & \textbf{0.97} & \textbf{0.97} && \textbf{0.99} & 0.98 & 0.98 && \textbf{0.93} & \textbf{0.93} & \textbf{0.93} \\
        & DEEPSVDD && \textbf{0.90} & 0.61 & 0.51 && \textbf{0.84} & 0.63 & 0.55 && \textbf{0.89} & 0.5 & 0.26 && \textbf{0.95} & 0.72 & 0.59 && \textbf{0.87} & 0.62 & 0.53 \\
        & NeuTraLAD && 0.86 & 0.87 & \textbf{0.94} && 0.90 & \textbf{0.93} & \textbf{0.93} && 0.86 & 0.91 & \textbf{0.96} && 0.90 & 0.95 & \textbf{0.97} && 0.88 & 0.90 & \textbf{0.94} \\
        & MemAE && \textbf{0.92} & \textbf{0.92} & \textbf{0.92} && \textbf{0.95} & 0.93 & 0.94 && \textbf{0.97} & 0.96 & 0.96 && \textbf{0.98} & \textbf{0.98} & \textbf{0.98} && \textbf{0.93} & 0.92 & \textbf{0.93} \\
      \bottomrule
    \end{tabular}
\end{table*}

\subsection{FedProx}\label{subsec:fedprox}

Aggregation strategies, such as FedProx \cite{fedprox}, are a critical axis of comparison for novel unsupervised anomaly detection methods in federated learning. To prove the comprehensiveness of this benchmark, we incorporated additional experiments utilizing the FedProx aggregation algorithm in a 3-client FL scenario. FedProx introduces a proximal term to the clients' local loss function, discouraging significant deviations from the global model. The modified loss function is defined as:

\begin{equation*}
    L = L_{obj} + \frac{\mu}{2}\|w - w_t\|^2
\end{equation*}

Here, $L_{obj}$ represents the reconstruction loss in the context of unsupervised anomaly detection, $w$ denotes the current local weights, $w_t$ indicates the global weights at communication round $t$, and $\mu$ is the coefficient of the proximal term.

By testing $\mu$ values in $\{0.01, 0.1, 1.0\}$, we observed that increasing $\mu$ often enhances the performance. This trend is particularly evident in the NeuTraLAD performance on the KDD10 dataset, indicating that FedProx effectively mitigates overfitting in this scenario. As discussed in \Cref{subsec:numclients}, the KDD10 dataset exhibits high homogeneity, leading to local overfitting, which FedProx's proximal term successfully addresses.

These findings show the ability of the FedProx algorithm in handling homogeneous datasets by alleviating local overfitting through its proximal term. This demonstrates the importance of selecting appropriate aggregation strategies to enhance the robustness and performance of federated learning models, particularly in the context of unsupervised anomaly detection.

\subsection{Metrics Unreliability}\label{subsec:unreliability}

\begin{table*}[t!]
    \centering
    \caption{\label{tab:FedProx}Applying FedProx in FL scenario with 3 clients and $\mu$ in $\{0.01,0.1,1.0\}$.}
    \begin{tabular}{l  l c ccc c ccc c ccc c ccc c ccc}
    \toprule
     \multirow{2}{*}{\rotatebox[origin=c]{90}{DS}}  & Method && \multicolumn{3}{c}{Precision} && \multicolumn{3}{c}{Recall} && \multicolumn{3}{c}{AUROC} && \multicolumn{3}{c}{AUPR}  && \multicolumn{3}{c}{F1}\\
     & $\mu$ &&   0.01 & 0.1 &  1.0 & &   0.01 & 0.1 &  1.0 & &   0.01 & 0.1 &  1.0 & &   0.01 & 0.1 &  1.0 & &   0.01 & 0.1 & 1.0 \\
     \midrule
     \multirow{5}{*}{\rotatebox[origin=c]{90}{Arrythmia}} 
        & DAE && 0.61 & \textbf{0.71} & 0.65 && 0.58 & 0.42 & \textbf{0.60} && 0.81 & 0.81 & \textbf{0.82} && 0.67 & \textbf{0.68} & \textbf{0.68} && 0.59 & 0.53 & \textbf{0.62} \\
        & DSEBM && 0.57 & \textbf{0.58} & 0.56 && 0.61 & \textbf{0.63} & \textbf{0.63} && \textbf{0.81} & \textbf{0.81} & \textbf{0.81} && \textbf{0.66} & \textbf{0.66} & \textbf{0.66} && 0.59 & \textbf{0.60} & 0.59 \\
        & DEEPSVDD && 0.44 & \textbf{0.51} & \textbf{0.51} && \textbf{0.67} & 0.37 & 0.56 && \textbf{0.68} & 0.65 & 0.65 && 0.54 & \textbf{0.55} & 0.53 && \textbf{0.53} & 0.43 &\textbf{ 0.53} \\
        & NeuTraLAD && 0.53 & 0.51 & \textbf{0.61} && \textbf{0.50} & 0.37 & 0.33 && 0.72 & 0.70 & \textbf{0.73} && \textbf{0.56} & 0.54 & 0.55 && \textbf{0.51} & 0.43 & 0.42 \\
        & MemAE && \textbf{0.72} & \textbf{0.72} & \textbf{0.72} && \textbf{0.40} & \textbf{0.40} &\textbf{ 0.40} && \textbf{0.81} & \textbf{0.81} & \textbf{0.81} && \textbf{0.66} & \textbf{0.66} & \textbf{0.66} && \textbf{0.52} & \textbf{0.52 }& \textbf{0.52} \\
    \midrule
      \multirow{5}{*}{\rotatebox[origin=c]{90}{Thyroid}} 
        & DAE && \textbf{0.22} & \textbf{0.22} & \textbf{0.22} && \textbf{0.43} & \textbf{0.43} & \textbf{0.43} && \textbf{0.86} & \textbf{0.86} & \textbf{0.86} && \textbf{0.28} & \textbf{0.28} & \textbf{0.28} && \textbf{0.29} & \textbf{0.29} & \textbf{0.29} \\
        & DSEBM && \textbf{0.23} & 0.22 & 0.22 && \textbf{0.16} & \textbf{0.16} & \textbf{0.16} && \textbf{0.74} & 0.73 & 0.73 && \textbf{0.19} & \textbf{0.19} & \textbf{0.19} && \textbf{0.19} & \textbf{0.19} & \textbf{0.19} \\
        & DEEPSVDD && 0.27 & 0.22 & \textbf{0.32} && 0.18 & \textbf{0.19} & 0.16 && 0.73 & 0.70 & \textbf{0.74} && 0.21 & 0.13 & \textbf{0.22} && \textbf{0.21} & 0.20 & \textbf{0.21} \\
        & NeuTraLAD && 0.63 & \textbf{0.65} & 0.58 && \textbf{0.80} & 0.68 & 0.78 && \textbf{0.97} & 0.95 & 0.96 && \textbf{0.69} & 0.67 & 0.68 && \textbf{0.71} & 0.66 & 0.67 \\
        & MemAE && \textbf{0.22} & \textbf{0.22} & \textbf{0.22} && \textbf{0.43} & \textbf{0.43} & \textbf{0.43} && \textbf{0.86} & \textbf{0.86} & \textbf{0.86} && \textbf{0.28} & \textbf{0.28} & \textbf{0.28} && \textbf{0.29} & \textbf{0.29} & \textbf{0.29} \\
    \midrule
      \multirow{5}{*}{\rotatebox[origin=c]{90}{KDD10}} 
        & DAE && \textbf{0.93} & \textbf{0.93} &\textbf{ 0.93} && \textbf{1.00} & \textbf{1.00} &\textbf{ 1.00} && \textbf{0.98} & \textbf{0.98} & \textbf{0.98} && \textbf{0.94} & \textbf{0.94} & \textbf{0.94} && \textbf{0.96} &\textbf{ 0.96} & \textbf{0.96} \\
        & DSEBM && \textbf{0.97} & \textbf{0.97} & \textbf{0.97} && \textbf{0.99} & \textbf{0.99} & \textbf{0.99} && \textbf{0.99} & \textbf{0.99} & \textbf{0.99} && \textbf{0.97} & \textbf{0.97} & \textbf{0.97} && \textbf{0.98} & \textbf{0.98} & \textbf{0.98} \\
        & DEEPSVDD && \textbf{0.88} & 0.73 & 0.69 && 0.74 & \textbf{0.84} & 0.78 && 0.79 & \textbf{0.88} & 0.83 && 0.83 & \textbf{0.87} & 0.82 && \textbf{0.81} & 0.78 & 0.73 \\
        & NeuTraLAD && 0.60 & 0.33 & \textbf{0.92} && 0.69 & 0.37 & \textbf{0.78} && 0.83 & 0.41 & \textbf{0.87} && 0.63 & 0.51 & \textbf{0.87} && 0.64 & 0.35 & \textbf{0.85 }\\
        & MemAE && \textbf{0.93} & 0.87 & 0.90 && \textbf{1.00} & 0.97 & \textbf{1.00} && \textbf{0.98} & \textbf{0.98} & \textbf{0.98} && \textbf{0.96} & 0.95 & 0.95 && \textbf{0.96} & 0.92 & 0.94 \\
        
    \midrule
      \multirow{5}{*}{\rotatebox[origin=c]{90}{NSL-KDD}} 
        & DAE && \textbf{0.93} & \textbf{0.93} & \textbf{0.93} && \textbf{0.94} & \textbf{0.94} & \textbf{0.94} && \textbf{0.97} & \textbf{0.97} & \textbf{0.97} && \textbf{0.98} & \textbf{0.98} & \textbf{0.98} && \textbf{0.93} & \textbf{0.93} & \textbf{0.93} \\
        & DSEBM && \textbf{0.95} & \textbf{0.95} & \textbf{0.95} && \textbf{0.92} & \textbf{0.92} & \textbf{0.92} && \textbf{0.97} & \textbf{0.97} & \textbf{0.97} && \textbf{0.99} & \textbf{0.99} & \textbf{0.99} &&\textbf{ 0.93} & \textbf{0.93} & \textbf{0.93} \\
        & DEEPSVDD && 0.86 & \textbf{0.89} & 0.80 && \textbf{0.94} & 0.93 & 0.74 && 0.91 & \textbf{0.94} & 0.72 && 0.95 & \textbf{0.97} & 0.86 && 0.90 & \textbf{0.91} & 0.77 \\
        & NeuTraLAD && 0.90 & \textbf{0.92} & 0.77 && 0.93 &\textbf{ 0.94} & 0.81 && \textbf{0.94} & \textbf{0.94} & 0.79 && \textbf{0.96} & \textbf{0.96} & 0.89 && 0.92 & \textbf{0.93} & 0.79 \\
        & MemAE && \textbf{0.92} & \textbf{0.92} & \textbf{0.92} && \textbf{0.95} & \textbf{0.95} & \textbf{0.95} && \textbf{0.97} & \textbf{0.97} & 0.97 && \textbf{0.98} & \textbf{0.98} & \textbf{0.98} && \textbf{0.93} & \textbf{0.93} & \textbf{0.93} \\
      \bottomrule
    \end{tabular}
\end{table*}

The reliability of evaluation metrics in anomaly detection is a critical issue that warrants closer examination. In this study, we focus on the overly optimistic nature of the AUROC metric.

The AUROC metric can present an overly optimistic assessment of model performance when considered in isolation. Although AUROC is a threshold-independent metric, it suffers from bias when evaluated across all possible thresholds. It considers both the true positive rate (TPR $ = \frac{TP}{TP + FN}$) and the false positive rate (FPR $= \frac{FP}{FP + TN}$) equally. In datasets where negative samples (normal cases) vastly outnumber positive samples (anomalies), the large number of true negatives (TN) dominates the denominator in the FPR calculation, thereby minimizing the FPR value and inflating the AUROC score.

While AUROC and F1-score are important metrics for anomaly detection, our key point is to highlight that these metrics can be biased and potentially manipulated \cite{boostf1}. To mitigate these issues, we recommend considering the Area Under the Precision-Recall Curve (AUPR) as a complementary metric. AUPR is threshold-independent and does not suffer from the same biases as AUROC, making it more suitable for evaluating anomaly detection applications. However, it is crucial to adopt a holistic approach by considering multiple metrics to provide a comprehensive evaluation of model performance. Metrics such as precision, recall, F1-score, AUROC, and AUPR should all be examined together to ensure a balanced and accurate assessment, avoiding the biases associated with relying on a single metric. We further introduced the stratified splitting of the test data into validation and test sets to avoid such manipulation of threshold-based metrics and to make sure all models are tested on equal grounds.

\section{Limitations}

We would like to shed the light on the limitation which pertains to the complexity of models used for unsupervised anomaly detection. These models often employ heuristics that are not propagated during the aggregation process during FL training. This leads to the loss of these heuristics at each round of communication, an example of this is the Memory module in MemAE which stores prototypical elements of the encoded normal data. Such limitation necessitates the development of specialized aggregation algorithms capable of meaningfully aggregating these heuristics on the server side. 

Another limitation is the continued use of certain datasets, such as KDD10, as benchmarks \cite{alvarez2022revealing}. These datasets have demonstrated lack of diversity, resulting in very high performance in most experiments, which does not provide meaningful insights. This calls attention to the need for increased research attention on tabular data, which \bmName aims to push forward.

\section{Conclusion and Future Work}
In this paper, we introduced \bmName, a benchmark framework for unsupervised anomaly detection in FL environments, addressing the unique challenges posed by decentralized data, model aggregation, and metric unreliability. We conducted extensive experiments comparing recent deep learning methods across multiple datasets and metrics, offering a comprehensive evaluation of their performance. Our results reveal that while centralized learning typically outperforms FL, federated setups can surpass centralized ones in specific scenarios due to their inherent regularization effects, which help mitigate overfitting. This highlights the potential of FL in enhancing the robustness of unsupervised anomaly detection models.

For future research, a more diverse collection of datasets from different domains need to be incorporated to enhance the generalizability of the findings. Additionally, aggregation strategies that are more suitable for unsupervised anomaly detection models are required seeking to improve the robustness and performance of FL models. Finally, experimenting with a wider range of models will help determine their effectiveness in unsupervised anomaly detection within FL frameworks.

\section*{Acknowledgment}

This work was supported by the BMWK project EuroDaT (Grant 68GX21010K).

\FloatBarrier
\bibliographystyle{splncs04} 
\bibliography{ref}

\end{document}